\documentclass[runningheads]{llncs}
\usepackage{graphicx}
\usepackage{comment}
\usepackage{amsmath,amssymb}
\usepackage{color}
\usepackage{url}
\usepackage{hyperref}

\usepackage{subcaption}

\begin{document}
\title{Teacher Network Calibration Improves Cross-Quality Knowledge Distillation}
%
%
\author{Pia \v{C}uk\inst{1}\inst{3} \and
Robin Senge\inst{2} \and
Mikko Lauri\inst{3}\and
Simone Frintrop\inst{3}}
\authorrunning{P. \v{C}uk et al.}
%
\institute{AdaLab UG, 20359 Hamburg, Germany\\
\email{pia.cuk@adalab.ai} \and
inovex GmbH, 76131 Karlsruhe, Germany\\
\email{rsenge@inovex.de} \and
Department of Computer Science, Universität Hamburg,\\
22527 Hamburg, Germany\\
\email{\{mikko.lauri, simone.frintrop\}@uni-hamburg.de}}
\maketitle              
\begin{abstract}
We investigate cross-quality knowledge distillation (CQKD), a knowledge distillation method where knowledge from a teacher network trained with full-resolution images is transferred to a student network that takes as input low-resolution images.
As image size is a deciding factor for the computational load of computer vision applications, CQKD notably reduces the requirements by only using the student network at inference time.
Our experimental results show that CQKD outperforms supervised learning in large-scale image classification problems.
We also highlight the importance of calibrating neural networks: 
we show that with higher temperature smoothing of the teacher's output distribution, the student distribution exhibits a higher entropy, which leads to both, a lower calibration error and a higher network accuracy. The implementation is available at: \url{https://github.com/PiaCuk/distillistic}.
\end{abstract}
\section{Introduction}
The increasing number of parameters in state-of-the-art deep neural networks (DNNs) in recent years \cite{menghani2021efficient} makes it challenging to deploy them in application areas such as robotics or embedded systems where only limited computational resources are available.
Knowledge distillation, or KD~\cite{bucilua2006model,hinton2015distilling}, is one of the techniques used to address the challenge to deploy DNNs in such applications.
KD transfers knowledge from a large, high-capacity teacher DNN to a smaller, computationally lightweight student DNN.
This is done by, for example, training the student to mimic the outputs of the teacher network.
It has been shown that KD can help the student network reach performance close to or equal to the performance of the teacher network, with a fraction of the computational cost at inference time \cite{romero2014fitnets,urban2016deep}.
Beyond its initial use for model compression, KD has proven successful as a technique to increase the accuracy of high-capacity DNNs \cite{xie2020self}, and has been used for pre-training for state-of-the-art models \cite{xie2020self}.

Building on the empirical success of KD, prior research also adapted the algorithm for knowledge transfer between domains. Cross-modal knowledge distillation, or CMKD, distills from a teacher with a different input modality than the student, e.g. from RGB to depth images \cite{gupta2016cross,tian2019contrastive}. 
Here, the student learns to mimic the feature space of the teacher. 
This has the advantage that the student's input modality can be unlabeled, as long as it is paired with an input from the teacher's modality.

In this paper, we investigate Cross-Quality Knowledge Distillation (CQKD), originally introduced in~\cite{DBLP:conf/bmvc/SuM17}, which is complementary to the CMKD methods mentioned above.
Instead of focusing on distinct modalities, such as RGB and depth images, the inputs of the teacher and student network are of the same modality, but different quality or resolution.
Specifically, the input of the teacher network is a high-resolution image, while the input of the student network is a downsampled, low-resolution variant of the same image.
CQKD allows the student network to distill knowledge from the teacher network, while additionally saving computational effort by processing lower-resolution input images.
We empirically evaluate CQKD on large-scale image classification on the ImageNet dataset~\cite{russakovsky2015imagenet}. 
Our results show that student networks trained via CQKD outperform baseline networks trained via standard supervised learning, while only incurring a less than 10\% increase in training time.

Additionally, our experimental results highlight the importance of proper calibration of the uncertainty estimates of DNNs~\cite{menon2021statistical}.
We find that it is essential for the teaching signal to reflect the uncertainty of the predictive task of the student. 
By applying stronger temperature smoothing to the teacher's output distribution, the student is trained to output a distribution with higher entropy. 
This appropriately reflects the uncertainty of the predictive task on low-resolution images. 
We show that student networks trained with a high-entropy teacher have a lower calibration error and a higher accuracy. These findings show the importance of calibration. 

The remainder of the paper is organized as follows.
In Section~\ref{sec:related}, we review related work in KD.
Section~\ref{sec:methods} reviews cross-quality knowledge distillation, which we empirically evaluate in Section~\ref{sec:experiments}.
Section~\ref{sec:conclusion} concludes the paper.
\section{Related Work}
\label{sec:related}
KD methods for DNNs can be categorized along multiple axes.
For instance, knowledge from the teacher network may be distilled to the student network via the output logits, e.g.,~\cite{DBLP:conf/nips/BaC14,hinton2015distilling}, or via intermediate layer activation maps, e.g.,~\cite{romero2014fitnets}.
There may be more than one teacher such as in~\cite{mirzadeh2020improved,zhang2018deep}, or an explicit teacher network may be replaced by an otherwise generated teaching signal~\cite{DBLP:conf/cvpr/YuanTLWF20}.
Offline KD methods such as~\cite{hinton2015distilling} first train a teacher network, and then distill its knowledge to a student network, while online KD methods such as~\cite{zhang2018deep} combine the two phases.

KD is practically relevant for deployment of neural networks in applications with computational limitations, and has therefore attracted significant research attention in recent years.
A complete survey of the area is outside the scope of this paper, and we instead refer the reader to recent surveys for a more complete overview of the literature~\cite{gou2021knowledge,DBLP:journals/pami/WangY22}.
In the following, we focus on two subareas of knowledge distillation relevant for our work: cross-modal and cross-quality distillation, and efforts to understand what makes for an effective teaching signal.

\paragraph{Cross-modal and cross-quality knowledge distillation.}
In cross-modal knowledge distillation (CMKD), the teacher and student networks' inputs are of a different modality.
Thus, the knowledge transferred from the teacher to the student needs to abstract from the input domain, i.e. it has to represent the content independently of modalities. 
Typically, the training data contains paired samples for both modalities. The student data is often not annotated, requiring a knowledge transfer from the teacher modality.
Gupta \textit{et al.}~\cite{gupta2016cross} transfer knowledge from RGB images to depth images and optical flow.
The teacher network is pre-trained on RGB images.
The student learns to replicate the feature representation of the teacher from a depth image or optical flow and therefore does not require training annotations. This distillation setup can be used as a pre-training for depth or motion vision tasks, where ground truth information is often very costly to acquire~\cite{dosovitskiy2015flownet}. 
Dai \textit{et al.}~\cite{dai2021learning} use CMKD from optical flow to RGB images for action detection in untrimmed video footage. 
Miech \textit{et al.}~\cite{miech2021thinking} employ distillation for multimodal transformer networks, that are trained with image-text pairs. Both teacher and student use both modalities, making this a special case of cross-modal distillation.
More examples for CMKD can be found in~\cite[Sect.~5.3]{gou2021knowledge}, especially for combinations of modalities that have been tested.

There are comparatively fewer works addressing the setting where the teacher and student networks process input images of different resolutions.
In~\cite{DBLP:conf/icip/PengHYS16}, the problem of fine-grained classification was considered.
However, no explicit teacher-student network separation was made.
Instead, the same network was first trained on high-resolution images, and later fine-tuned on low-resolution images.
The empirical results confirmed that knowledge of high-resolution features can be helpful for low-resolution classification tasks.
Su and Maji~\cite{DBLP:conf/bmvc/SuM17} propose cross-quality knowledge distillation (CQKD) where both the teacher and student networks are pre-trained on the ImageNet data.
They find cross-quality distillation performs better than simply fine-tuning the student network on low-resolution images or training the student first on high-resolution and then on low-resolution data in two stages.
We demonstrate our method on the larger-scale ImageNet classification task, in contrast to the smaller datasets of around 10k samples each in~\cite{DBLP:conf/bmvc/SuM17}.
Furthermore, we additionally analyze the importance of calibration for cross-quality distillation.
Finally, in our experiments we do not use pre-trained weights for the student network.

\paragraph{What makes for an effective teaching signal?}
Knowledge distillation is partly motivated by human learning, where a teacher can help a student learn more effectively by providing feedback beyond the correctness of the student's answer.
However, few works investigate the mathematical foundations of KD to understand the mechanics of the algorithm and what type of teaching signal would be most effective. Nevertheless, this is essential to, e.g. explain phenomena such as the capacity gap that hinders lightweight student networks from learning effectively from large, highly accurate DNNs~\cite{mirzadeh2020improved}.

Recent research has investigated the question of what kind of teaching signal is most effective for KD from complementary perspectives: either by a theoretical analysis of the learning task~\cite{menon2021statistical} or empirically by omitting a teacher network entirely~\cite{DBLP:conf/cvpr/YuanTLWF20} and replacing it with a manually designed teaching signal.
Menon \textit{et al.}~\cite{menon2021statistical} prove that a teacher that accurately approximates the true Bayes class probabilities $\mathbb{P}(y \mid x)$ in its output can lower the variance of the student's and thus improve performance.
How well a predictor such as a DNN approximates the Bayes class probabilities is quantified by \emph{calibration}.
Despite being accurate, modern DNNs are susceptible to poor calibration~\cite{guo2017calibration}.
As argued in~\cite{menon2021statistical}, a poorly calibrated network makes for a poor teacher.
In~\cite{DBLP:conf/cvpr/YuanTLWF20} the teacher network is replaced by a teaching signal obtained by adapting \emph{label smoothing}~\cite{DBLP:conf/cvpr/SzegedyVISW16}, a well known regularization technique for reducing the overconfidence of deep neural networks.
This empirical method for reducing overconfidence, and thereby improving calibration, is also shown to be effective for KD.

Based on these theoretical and empirical findings, we conclude that it is important to track the quality of the teaching signal by metrics such as the expected calibration error (ECE,~\cite{naeini2015obtaining}).
In our experiments, we measure the ECE to understand the performance of CQKD and take measures for re-calibration when needed.

\begin{figure}[t]
    \centering
    \includegraphics[width=0.99\textwidth]{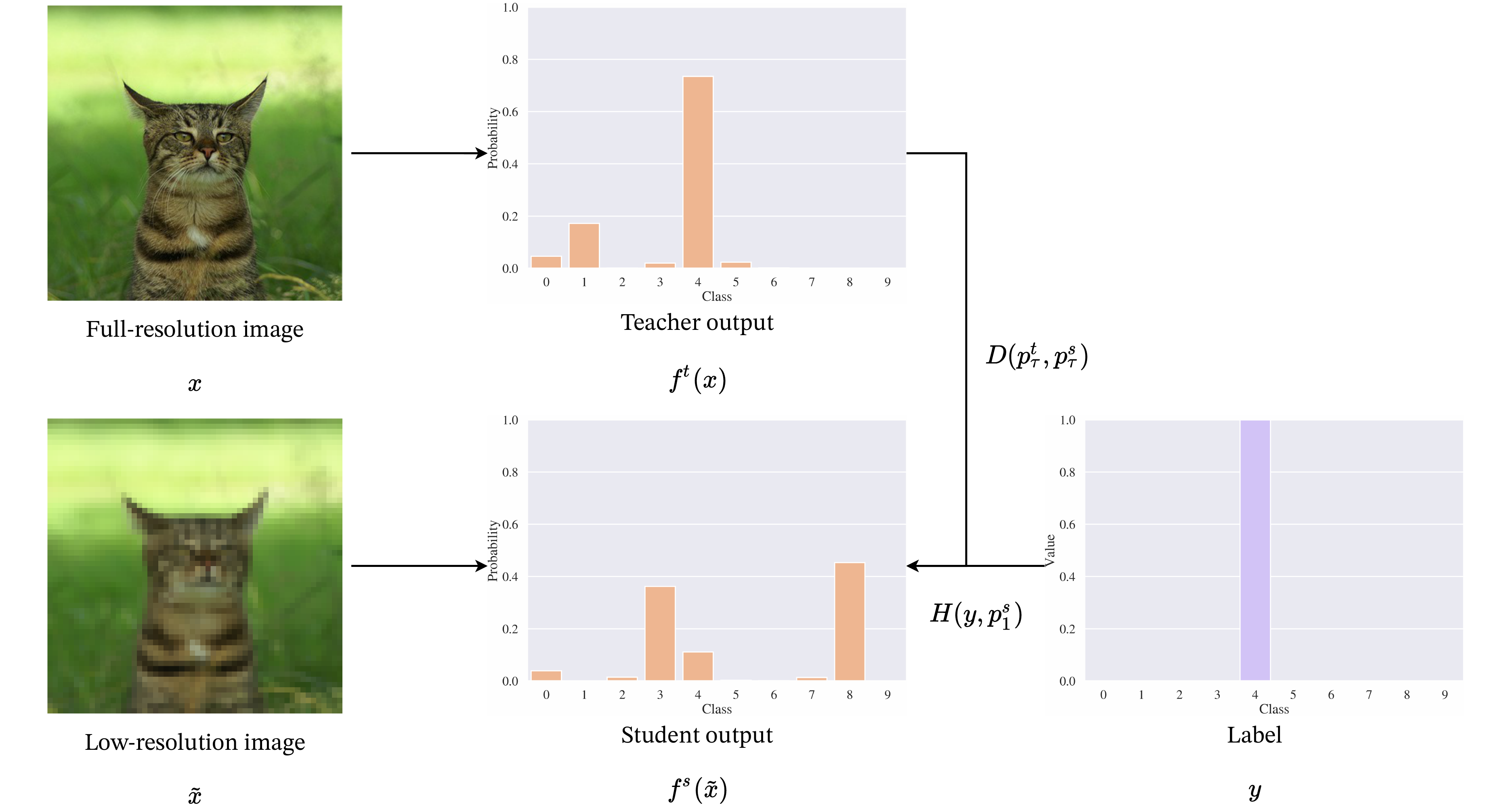}
    \caption[Training pipeline for cross-quality distillation.]{In cross-quality knowledge distillation (CQKD), the input to the teacher network is a full-resolution image, and the student receives a downscaled low-resolution image as input. The student is trained with a weighted loss using the cross-entropy on the target label and the Kullback-Leibler divergence between the student and teacher output distributions. At inference time, the student network is retained and only low-resolution images need to be processed.}
    \label{fig:cross-quality_kd}
\end{figure}

\section{Cross-Quality Knowledge Distillation}
\label{sec:methods}
We now review Cross-Quality Knowledge Distillation (CQKD)~\cite{DBLP:conf/bmvc/SuM17}, where the student networks learns to predict on low-resolution images, using the target label and the output of the high-capacity teacher as input. 
The teacher network receives the full-resolution images as input, which is why the knowledge is transferred across different image qualities.
Similar to the KD method of~\cite{hinton2015distilling}, CQKD is also an offline distillation method where we first train the teacher network via supervised learning, and then transfer the knowledge to a student network in a distillation phase.
After the distillation phase, the teacher network can be discarded, and at inference time only the student network processing low-resolution images is used.
A schematic overview of the distillation phase of CQKD is shown in Figure~\ref{fig:cross-quality_kd}.
We next describe the distillation phase for the image classification task we target.

We are given a full resolution input image $x$ along with its target label $y \in \{1, 2, \ldots, K\}$.
By feeding the image $x$ to the teacher network $f^t$, we obtain the vector of teacher logits $\mathbf{z}^t = f^t(x)$.
Next, we create a low-resolution version $\tilde{x}$ by downsampling $x$ by a constant factor\footnote{We evaluate the effect of the downsampling factor in the next section.}.
The low-resolution image is fed to the student network $f^s$, resulting in the student logits $\mathbf{z}^s = f^s(\tilde{x})$.

For later use, we will also define temperature smoothing or soft targets~\cite{hinton2015distilling}.
Temperature smoothing re-calibrates predictive models by distributing the probability mass between classes more uniformly~\cite{guo2017calibration}.
Soft targets are also one of the essential components for the successful transfer of knowledge with KD~\cite{DBLP:conf/cvpr/YuanTLWF20,menon2021statistical}.
Given a vector $\mathbf{z}^t = \begin{bmatrix}z_1^t & z_2^t & \ldots & z_K^t\end{bmatrix}$ of teacher logits and a temperature hyperparameter $\tau \in \mathbb{R}$, the $i$th soft target is defined as

\begin{equation}
  p^t_\tau(z_i^t) =  \frac{\exp (\frac{z_i^t}{\tau})} {\sum_j \exp (\frac{z_j^t}{\tau})}.
  \label{eq:soft_targets}
\end{equation}

The definition is analogous for the student logits $\mathbf{z}^s$.
We denote the entire vector of teacher soft targets and student soft targets by $p_\tau^t$ and $p_\tau^s$, respectively.
Note that setting $\tau=1$ recovers the Softmax activation function commonly applied in classification networks.

To train the student network, we apply a loss function defined as a weighted sum of a label loss term that compares the student logits with the target label, and a distillation loss term that compares the student logits with the teacher logits.
The CQKD loss is otherwise similar to the KD loss of~\cite{hinton2015distilling}, except that the teacher and student logits originate from images of different resolutions.
The CQKD loss to be minimized is defined as

\begin{equation}
  \mathcal{L}_\text{CQKD}  = (1-\alpha) H(y,p^s_1) + \alpha D(p_{\tau}^t, p_{\tau}^s),
  \label{eq:L_KD}
\end{equation}

where $\alpha \in (0,1)$ is a weighting factor, $H(y,p^s_1)$ is the cross-entropy between the target label and the student output without temperature smoothing, and $D(p_{\tau}^t, p_{\tau}^s)$ is the Kullback-Leibler divergence between the soft targets of teacher and student.
By choosing the weighting factor, we can change the balance between training the student network to accurately predict the target label and training it to mimic the teacher's output distribution. In our experiments, we use $\alpha = 0.5$ throughout.
\section{Empirical Evaluation}
\label{sec:experiments}
We empirically compare CQKD to a supervised learning baseline, and an alternative online KD algorithm.
We evaluate the effect of downsampling and temperature smoothing on the performance of the student network, and quantitatively compare the student networks in terms of \emph{calibration}, the ability of the networks to output accurate probability estimates reflecting the true uncertainty in their predictions.
In the next subsections, we review the experimental setup and evaluation metrics, present the results of the experiments, and then discuss the significance of the results. The implementation is available at: \url{https://github.com/PiaCuk/distillistic}.

\subsection{Experimental setup}
For cross-quality distillation, the teacher always receives the input images with dimensions $224 \times 224$ (full resolution), while the students receive the images downscaled. 
We experiment with downscaled images of size $h \times h$, with $h$ equal to 168, 112, 56, 42, or 28.
For CQKD, we tested soft targets with $\tau = 10$ and $\tau = 20$ in the divergence loss.

\paragraph{Dataset and network architectures.}
In all our experiments, we use the ImageNet~\cite{russakovsky2015imagenet}, a 1000-class image classification dataset.
ImageNet is particularly suitable for cross-quality distillation, as all images are at least $224 \times 224$ pixels in width and height, allowing for sufficient opportunity for downsampling. 
We use a pre-trained ResNet-50 teacher network from \emph{torchvision}~\cite{paszke2019pytorch}, scoring 76.130\% top-1 and 92.862\% top-5 accuracy on the ImageNet full-resolution validation set.
All student networks we use are ResNet-18 networks~\cite{he2016deep}, that are initialized randomly.

\paragraph{Baselines.}
We compare CQKD to two main baselines.
First, we consider a supervised learning baseline where a ResNet-18 student model is trained directly with the cross-entropy loss on low-resolution images.
Second, we consider deep mutual learning (DML)~\cite{zhang2018deep}, an online knowledge distillation method with a cohort of ResNet-18 student networks trained concurrently on the low-resolution images.
We choose DML as it has previously obtained good performance on ImageNet~\cite{zhang2018deep}, showing significant improvements over the supervised learning baseline.
In each training step of DML, we iterate over the student networks $i=1, 2, \ldots, m$, training the $i$th network using the loss function
\begin{equation}
  \mathcal{L}_{\text{DML},i} = H(y,p^i) + \frac{1}{m-1} \sum_{j=1, j \neq i}^m D(p^j, p^i),
  \label{eq:L_DML}
\end{equation}
where $p^i$ and $p^j$ are the output probabilities of the $i$th and $j$th student network, respectively.
We use a student cohort of size $m = 3$.

We do not evaluate on baselines such as directly training the student on low-resolution images, or stage-wise training on both datasets, as these were already considered in~\cite{DBLP:conf/bmvc/SuM17} and found to perform worse than CQKD.

\paragraph{Hyperparameter settings and implementation details.}
We use a cyclical learning rate schedule~\cite{smith2019super} with a maximum learning rate of $\eta = 0.001$ and an AdamW optimizer~\cite{loshchilov2017decoupled}. 
For all methods, we use RandAugment~\cite{cubuk2020randaugment} for data augmentation with two augmentation transformations per image and a magnitude of 9.
For each setting, we train for 20 epochs.
The experiments were conducted using an NVIDIA T4 GPU with CUDA 11.3.
\subsection{Evaluation metrics}
We record the average classification accuracy to measure the fraction of validation images correctly classified by the network.
For classification, it is often useful to not only predict a class but also include a measure of \textit{confidence}.
Confidence is the probability of a prediction being correct and accounts for uncertainty in the predictions. 
In the context of neural networks, the output of a model is scaled to have the characteristics of a probability distribution. 
However, this scaling does not automatically guarantee that the scores, which express the model's confidence, represent the accuracy of the predictions.
DNNs in particular are known to be prone to overconfident predictions~\cite{guo2017calibration}.
It is possible for a network to have high or low accuracy independently of how confident it is in its predictions.
Therefore, it is necessary to consider additional metrics other than accuracy to measure confidence.

We record the average entropy of the output distribution as a measure of the uncertainty related to the predictions of the network.
A network with low average entropy is confident in its predictions, whereas a network with high entropy is uncertain.
Finally, we measure the expected calibration error (ECE)~\cite{naeini2015obtaining} of the student networks.
ECE is calculated on the entire set of predictions output by the network as follows.
We first partition the interval $[0,1]$ of probabilities into $B$ equally spaced bins.
Then, we consider each output prediction of the network as a tuple $(\hat{y}, \hat{p}, y)$, where $\hat{y}$ is the predicted class, $\hat{p}$ is the probability associated with the predicted class, and $y$ is the true target label.
Each prediction of the network is assigned to the bin $b$ that covers $\hat{p}$.
ECE is calculated as
\begin{equation}
    \text{ECE} = \sum_{b=1}^B \frac{m_b}{n} \left| \texttt{acc}(b) - \texttt{conf}(b) \right|,
    \label{eq:ece}
\end{equation}
where $n$ is the total number of predictions, $m_b$ is the number of predictions in bin $b$, $\texttt{acc}(b)$ is the average accuracy of predictions in bin $b$ obtained from $\hat{y}$ and $y$, and $\texttt{conf}(b)$ is the average of the output probabilities $\hat{p}$ for the predictions in bin $b$.
The lower the ECE is, the better calibrated the network is.

\subsection{Results}

\begin{figure}[t]
    \centering
    \begin{subfigure}[t]{0.49\linewidth}
        \includegraphics[width=0.99\textwidth]{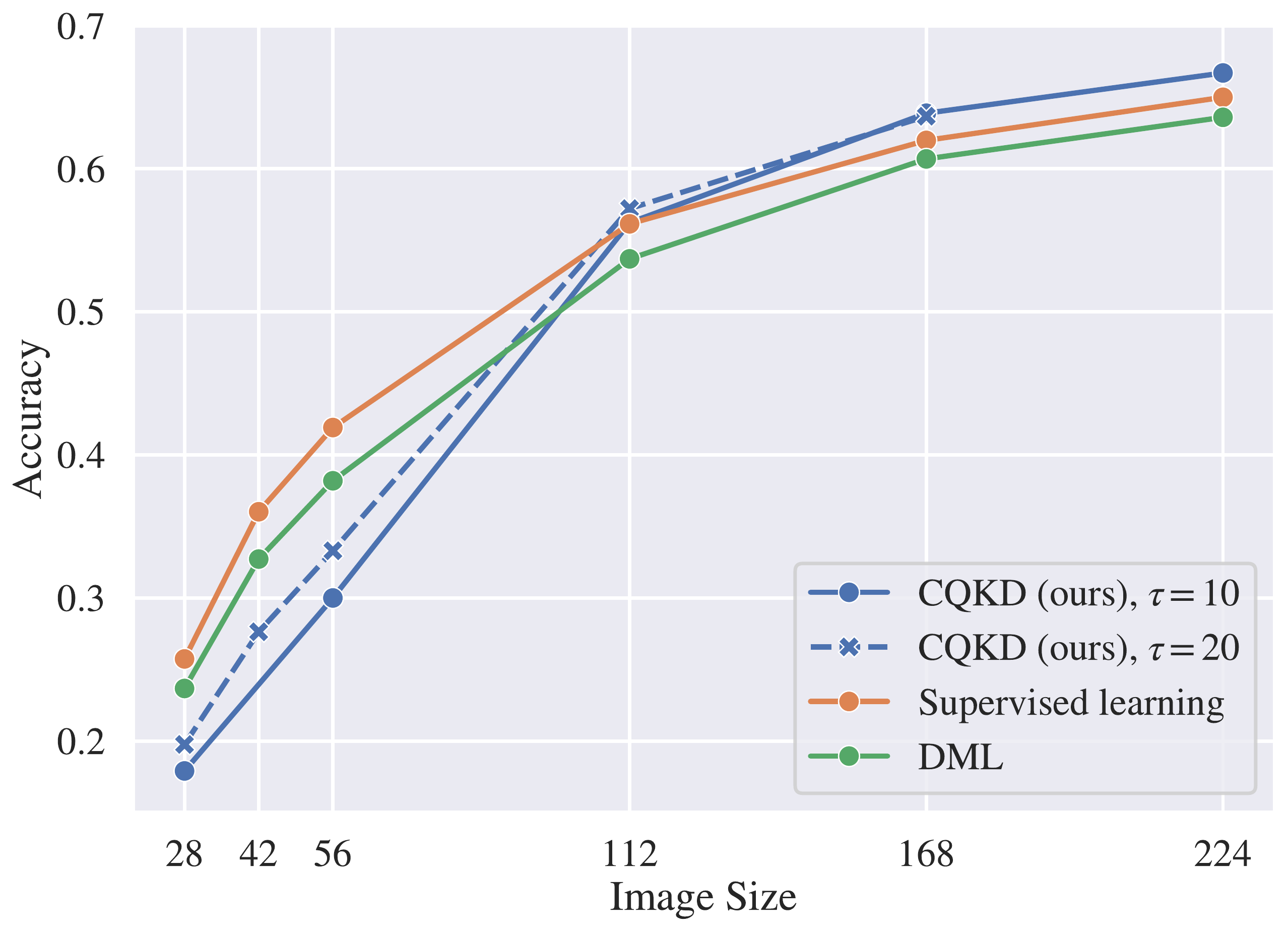}
        \caption{Validation accuracy.}
        \label{fig:downscaling_acc}
    \end{subfigure}
    \hfill
    \begin{subfigure}[t]{0.49\linewidth}
        \includegraphics[width=0.99\textwidth]{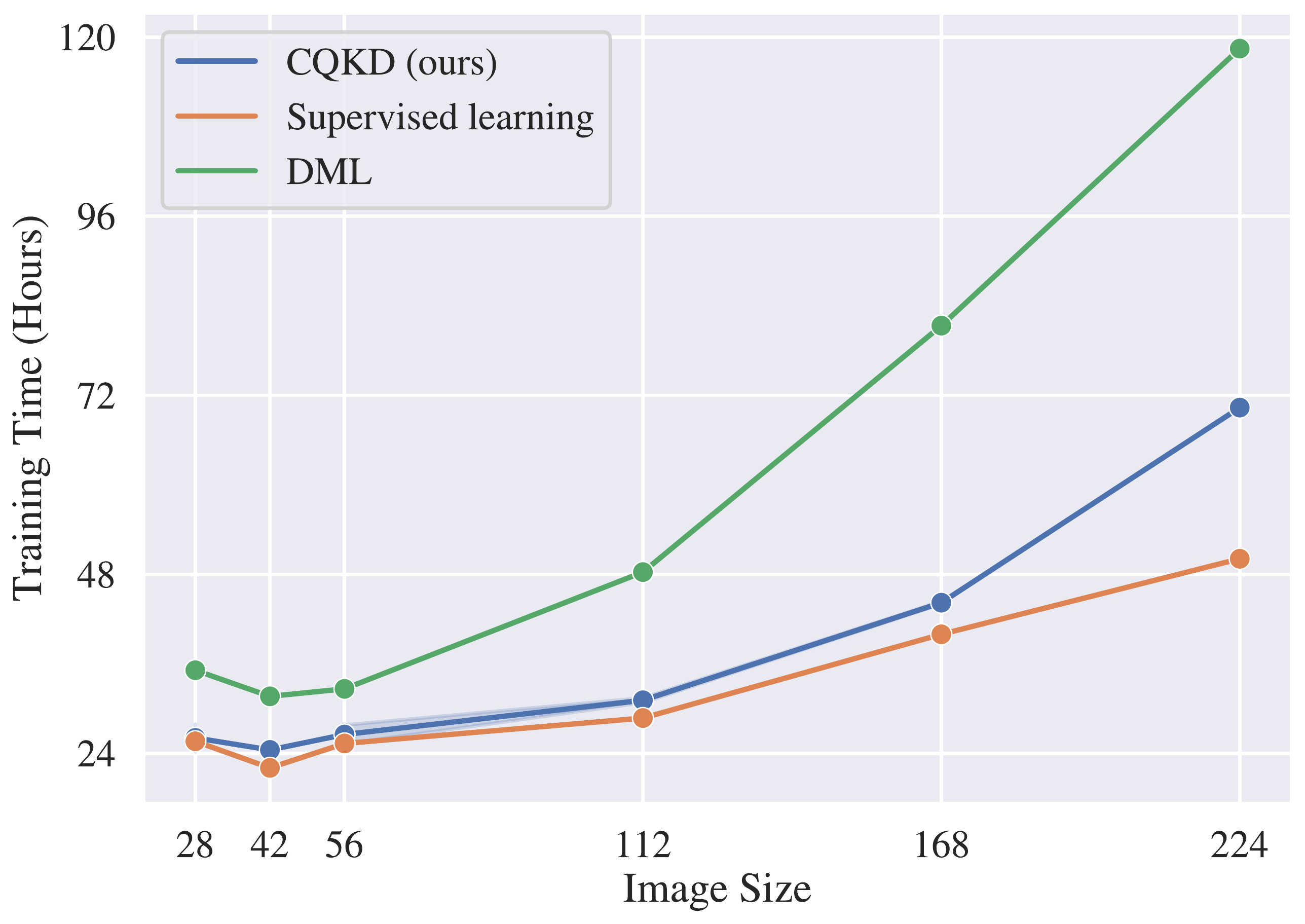}
        \caption{Training time.}
        \label{fig:downscaling_time}
    \end{subfigure}
    \vfill
    \begin{subfigure}[t]{0.49\linewidth}
        \includegraphics[width=0.99\textwidth]{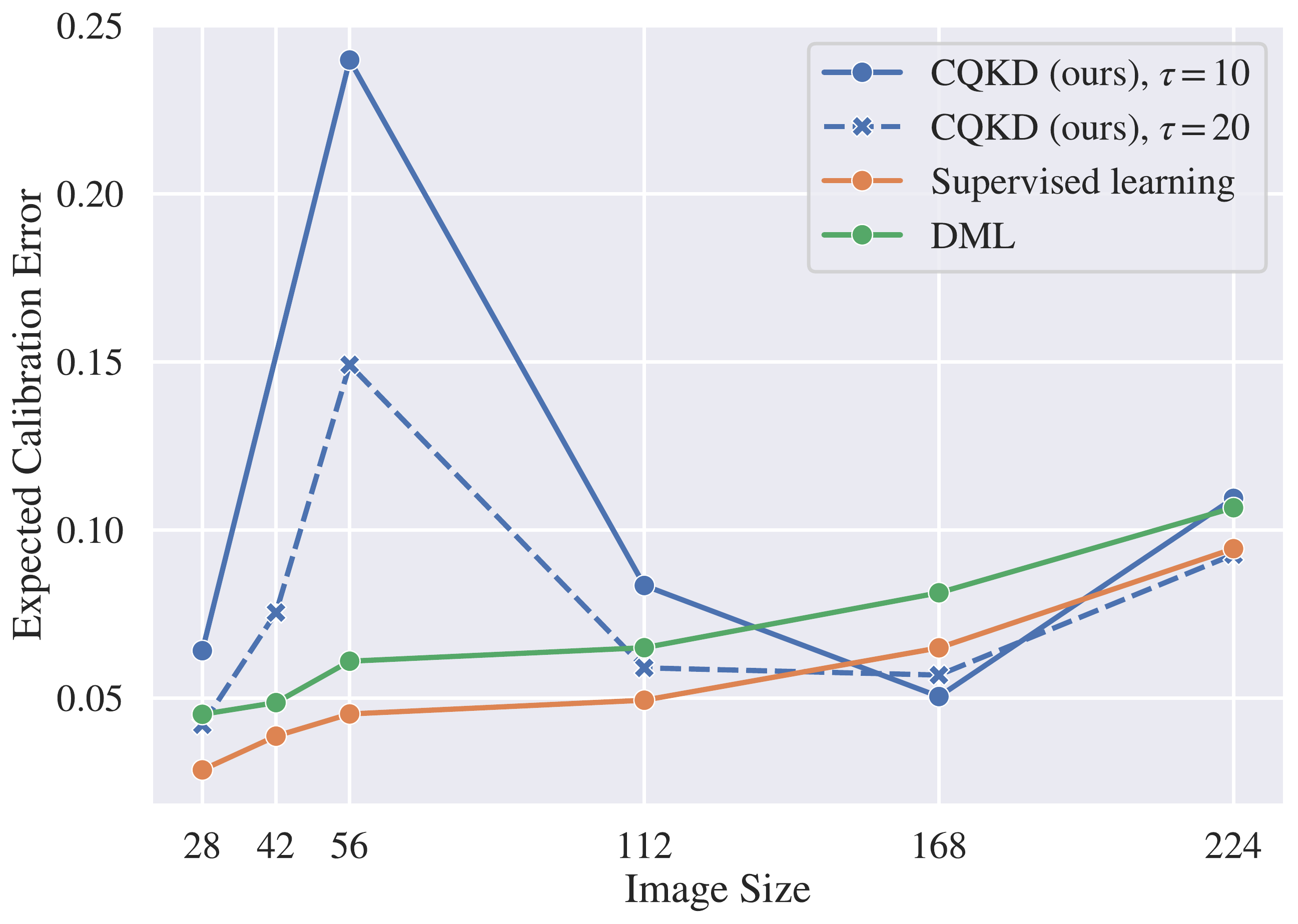}
        \caption{Expected calibration error (ECE).}
        \label{fig:downscaling_ece}
    \end{subfigure}
    \hfill
    \begin{subfigure}[t]{0.49\linewidth}
        \includegraphics[width=0.99\textwidth]{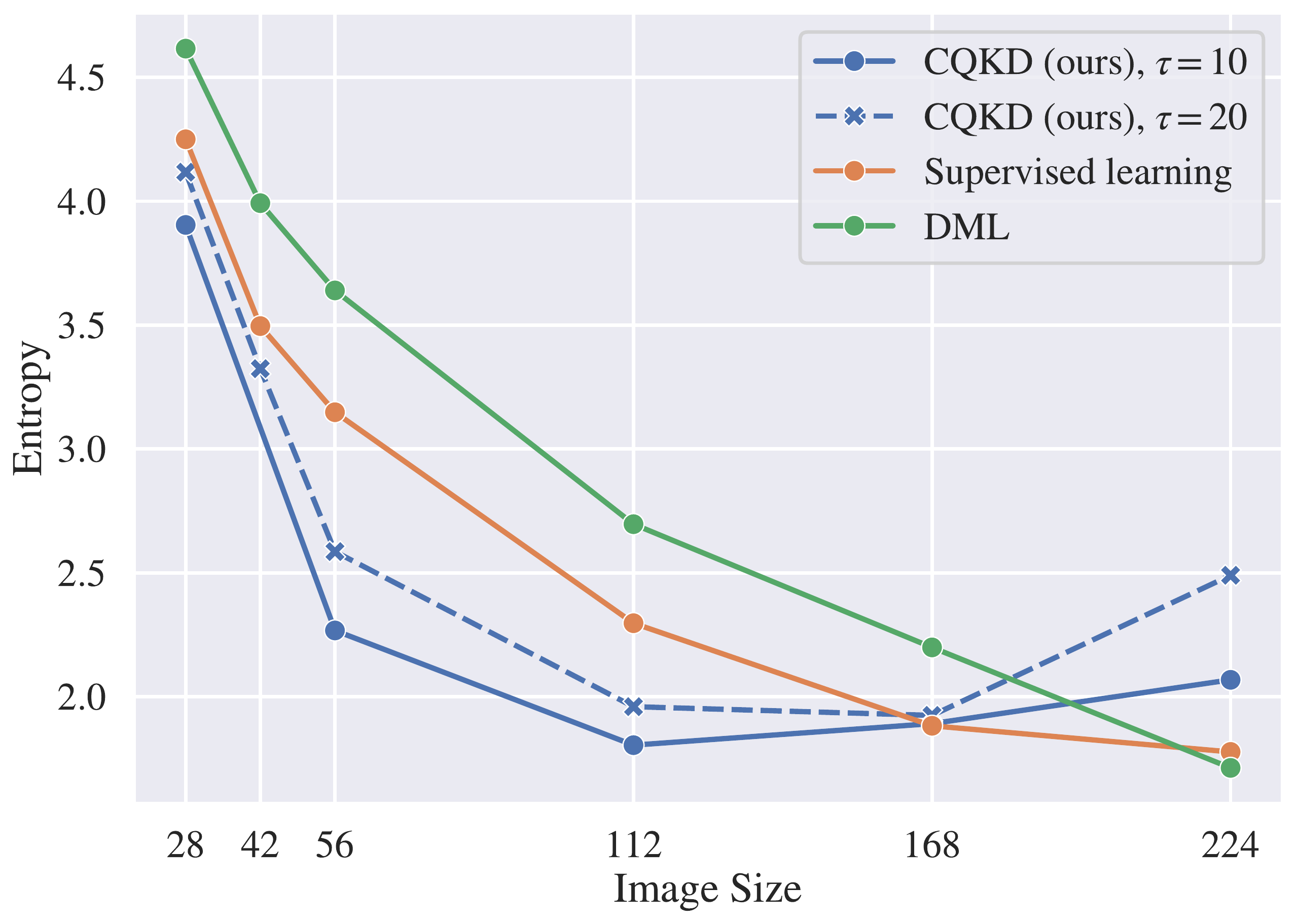}
        \caption{Entropy of the output distribution.}
        \label{fig:downscaling_entropy}
    \end{subfigure}
    \caption[Accuracy, training time, ECE, and entropy on ImageNet as a function of training image size.]{Accuracy, training time, ECE, and entropy on ImageNet as a function of training image size. For cross-quality KD, we report a setting with $\tau=10$ and $\tau=20$. Up to image size 112, the baseline and DML outperform KD. For larger image sizes, KD performs best.}
    \label{fig:downscaling}
\end{figure}

Figure \ref{fig:downscaling} shows the validation accuracy, training time, expected calibration error (ECE), and output distribution entropy as a function of the size $h$ of the student's input image $\tilde{x}$.
In terms of accuracy, CQKD outperforms both supervised learning and DML for an image size of 112 pixels and greater.
The accuracy of the supervised learning baseline and DML are similar, with DML scoring 2 to 3 percentage points lower.
As expected, the overall accuracy increases with higher resolution.

For image size smaller than 112, CQKD has a lower accuracy than supervised learning and DML (Fig.~\ref{fig:downscaling_acc}).
In these cases, the student image $\tilde{x}$ is less than one fourth of the size of the teacher image $x$.
We suspect that a large difference in image sizes makes it challenging to transfer knowledge from the teacher to the student, as the features the teacher and student networks can potentially learn also differ.
For image sizes above 112, there is no performance difference for CQKD with $\tau=10$ or $\tau=20$.
In contrast, CQKD with $\tau = 20$ outperforms $\tau = 10$ in accuracy for image sizes less than 112 pixels.
Setting $\tau=20$ corresponds to stronger smoothing and therefore lower confidence for the teaching signal, which helps the CQKD student avoid overconfident and inaccurate predictions.
This is also confirmed by the lower ECE (Fig.~\ref{fig:downscaling_ece}) for $\tau=20$ for image sizes less than and including 112. 

ECE (Fig.~\ref{fig:downscaling_ece}) measures how well the student network is calibrated, with lower values being better.
We note that CQKD with $\tau=20$ is the best calibrated with full resolution images (size 224), while also reaching the highest accuracy.
ECE alone does not convey whether the models are over- or under-confident in their predictions, which is why it is essential to evaluate it in combination with the entropy of the output distribution (Fig.~\ref{fig:downscaling_entropy}).
Informally speaking, entropy indicates whether the predicted distribution is closer to uniform (high entropy) or peaked (low entropy).
For image size 56, where CQKD has the highest ECE, both smoothing settings $\tau$ have a much lower entropy than the supervised learning baseline and DML. 
These results suggest that the CQKD models have poor ECE due to over-confidence, assigning a higher probability to the predicted class than the actual accuracy of the model.

For the supervised learning baseline and DML, ECE increases as a function of the student image size (Fig.~\ref{fig:downscaling_ece}).
As the image size increases and accuracy improves, entropy decreases indicating overconfidence, and the calibration of the student networks degrades.

On top of the time required to train the teacher model, CQKD requires only up to 10\% more training time than the supervised learning baseline (Fig. \ref{fig:downscaling_time}), while reaching a higher accuracy for image sizes above 112.
DML requires up to 2 times longer to train than the supervised learning baseline. This is not surprising, as in DML three networks instead of one are trained. 

\subsection{Discussion}
\label{subsec:discussion}
When using a teacher model trained on full-resolution images to transfer knowledge to a student network using low-resolution images, it is crucial to ``bridge the gap'' between the learning task of the student and the teacher's predictions.
The miscalibration of the CQKD student for small image sizes illustrates how the teaching signal biases the student to learn a solution that does not fit the problem.
The student learns to predict a low-entropy distribution that does not match the uncertainty inherent in the problem due to the decreased input image size. 
We quickly obtain a more fitting solution by increasing temperature smoothing of the teaching signal, showing how effective the simple method of soft targets is.
We believe that the key to making cross-quality distillation widely applicable is to find an effective, automatic way to adapt the teacher's distribution to the student's learning problem. 
Reflecting the aleatoric uncertainty of the problem through calibrated predictions is one fundamental component of it. 

As confirmed by our results, reducing the image resolution causes a degradation in accuracy in the student models. 
However, this is a trade-off worth considering for many applications. 
For CQKD and DML, reducing the resolution to 168 in width and height reduces the training time by a third while reducing the accuracy by only 5\%. 
This speed-up is especially interesting for CQKD, as it outperforms the baseline by almost two percentage points for the same resolution and scores only one percentage point lower than the baseline trained on full resolution. 
The trade-off between accuracy and computational efficiency is likely to be domain-dependent, and there may be applications where it is even more effective to use downscaling.

It is possible that slight changes in the overall results would still be observed when training for a greater number of epochs.
Especially the calibration of a model's predictions can change towards full convergence, as networks often become overconfident to match the target labels~\cite{guo2017calibration}. 
Nevertheless, we believe that the trends and differences between the algorithms remain unchanged, albeit at a lower accuracy.
\section{Conclusion}
\label{sec:conclusion}
We investigate cross-quality knowledge distillation (CQKD)~\cite{DBLP:conf/bmvc/SuM17} where knowledge from a teacher network trained with full-resolution images is transferred to a student network with low-resolution inputs.
We apply CQKD to large-scale image classification, and find that compared to other KD methods, CQKD obtains computational savings at inference time as the student network only processes a low resolution image.
CQKD outperforms supervised learning and an online knowledge distillation approach in large-scale image classification in terms of accuracy, while only incurring a modest increase in training time.

Our empirical results confirm the importance of measuring the calibration of the networks.
In particular, we found that it is crucial to apply re-calibration methods such as temperature smoothing to the teaching signal to avoid overconfident student networks when training with CQKD.
Applying temperature smoothing to the teacher's output distribution, we found that the student distribution has a greater entropy, lower calibration error, and a higher accuracy.

Exploring CQKD in other domains and problems is an exciting direction for future work.
This could lead to further insights about when the trade-off between lower accuracy and shorter training and inference times is warranted.
A theoretical analysis of CQKD could lead to a better understanding of the limitations of distillation when the student network obtains a compressed version of the teacher input.
This could be the first step towards automatic re-calibration methods that adapt the teacher's output distribution to the learning problem faced by the student network, further improving the applicability of CQKD.
%
%
%
\bibliographystyle{splncs04}
\bibliography{egbib}

\begin{thebibliography}{10}
\providecommand{\url}[1]{\texttt{#1}}
\providecommand{\urlprefix}{URL }
\providecommand{\doi}[1]{https://doi.org/#1}

\bibitem{DBLP:conf/nips/BaC14}
Ba, J., Caruana, R.: Do deep nets really need to be deep? In: Advances in
  Neural Information Processing Systems 27. pp. 2654--2662 (2014)

\bibitem{bucilua2006model}
Buciluǎ, C., Caruana, R., Niculescu-Mizil, A.: Model compression. In: ACM
  SIGKDD International Conference on Knowledge Discovery and Data Mining. pp.
  535--541 (2006)

\bibitem{cubuk2020randaugment}
Cubuk, E.D., Zoph, B., Shlens, J., Le, Q.V.: Randaugment: Practical automated
  data augmentation with a reduced search space. In: IEEE/CVF Conference on
  Computer Vision and Pattern Recognition Workshops. pp. 702--703 (2020)

\bibitem{dai2021learning}
Dai, R., Das, S., Bremond, F.: Learning an augmented rgb representation with
  cross-modal knowledge distillation for action detection. In: IEEE/CVF
  International Conference on Computer Vision. pp. 13053--13064 (2021)

\bibitem{dosovitskiy2015flownet}
Dosovitskiy, A., Fischer, P., Ilg, E., Hausser, P., Hazirbas, C., Golkov, V.,
  Van Der~Smagt, P., Cremers, D., Brox, T.: Flownet: Learning optical flow with
  convolutional networks. In: IEEE International Conference on Computer Vision.
  pp. 2758--2766 (2015)

\bibitem{gou2021knowledge}
Gou, J., Yu, B., Maybank, S.J., Tao, D.: Knowledge distillation: A survey.
  International Journal of Computer Vision  \textbf{129}(6),  1789--1819 (2021)

\bibitem{guo2017calibration}
Guo, C., Pleiss, G., Sun, Y., Weinberger, K.Q.: On calibration of modern neural
  networks. In: International Conference on Machine Learning. pp. 1321--1330
  (2017)

\bibitem{gupta2016cross}
Gupta, S., Hoffman, J., Malik, J.: Cross modal distillation for supervision
  transfer. In: {IEEE/CVF} Conference on Computer Vision and Pattern
  Recognition. pp. 2827--2836 (2016)

\bibitem{he2016deep}
He, K., Zhang, X., Ren, S., Sun, J.: Deep residual learning for image
  recognition. In: {IEEE/CVF} Conference on Computer Vision and Pattern
  Recognition. pp. 770--778 (2016)

\bibitem{hinton2015distilling}
Hinton, G., Vinyals, O., Dean, J.: Distilling the knowledge in a neural
  network. arXiv preprint arXiv:1503.02531  (2015)

\bibitem{loshchilov2017decoupled}
Loshchilov, I., Hutter, F.: Decoupled weight decay regularization. arXiv
  preprint arXiv:1711.05101  (2017)

\bibitem{menghani2021efficient}
Menghani, G.: Efficient deep learning: A survey on making deep learning models
  smaller, faster, and better. arXiv preprint arXiv:2106.08962  (2021)

\bibitem{menon2021statistical}
Menon, A.K., Rawat, A.S., Reddi, S., Kim, S., Kumar, S.: A statistical
  perspective on distillation. In: International Conference on Machine
  Learning. pp. 7632--7642 (2021)

\bibitem{miech2021thinking}
Miech, A., Alayrac, J.B., Laptev, I., Sivic, J., Zisserman, A.: Thinking fast
  and slow: Efficient text-to-visual retrieval with transformers. In:
  {IEEE/CVF} Conference on Computer Vision and Pattern Recognition. pp.
  9826--9836 (2021)

\bibitem{mirzadeh2020improved}
Mirzadeh, S.I., Farajtabar, M., Li, A., Levine, N., Matsukawa, A., Ghasemzadeh,
  H.: Improved knowledge distillation via teacher assistant. In: AAAI
  Conference on Artificial Intelligence. pp. 5191--5198 (2020)

\bibitem{naeini2015obtaining}
Naeini, M.P., Cooper, G., Hauskrecht, M.: Obtaining well calibrated
  probabilities using {Bayesian} binning. In: AAAI Conference on Artificial
  Intelligence (2015)

\bibitem{paszke2019pytorch}
Paszke, A., Gross, S., Massa, F., Lerer, A., Bradbury, J., Chanan, G., Killeen,
  T., Lin, Z., Gimelshein, N., Antiga, L., Desmaison, A., K{\"{o}}pf, A., Yang,
  E.Z., DeVito, Z., Raison, M., Tejani, A., Chilamkurthy, S., Steiner, B.,
  Fang, L., Bai, J., Chintala, S.: Pytorch: An imperative style,
  high-performance deep learning library. In: Advances in Neural Information
  Processing Systems 32. pp. 8024--8035 (2019)

\bibitem{DBLP:conf/icip/PengHYS16}
Peng, X., Hoffman, J., Yu, S.X., Saenko, K.: Fine-to-coarse knowledge transfer
  for low-res image classification. In: {IEEE} International Conference on
  Image Processing. pp. 3683--3687 (2016)

\bibitem{romero2014fitnets}
Romero, A., Ballas, N., Kahou, S.E., Chassang, A., Gatta, C., Bengio, Y.:
  Fitnets: Hints for thin deep nets. In: International Conference on Learning
  Representations (2015)

\bibitem{russakovsky2015imagenet}
Russakovsky, O., Deng, J., Su, H., Krause, J., Satheesh, S., Ma, S., Huang, Z.,
  Karpathy, A., Khosla, A., Bernstein, M., et~al.: Imagenet large scale visual
  recognition challenge. International journal of computer vision
  \textbf{115}(3),  211--252 (2015)

\bibitem{smith2019super}
Smith, L.N., Topin, N.: Super-convergence: Very fast training of neural
  networks using large learning rates. In: Artificial Intelligence and Machine
  Learning for Multi-domain Operations Applications. vol. 11006, pp. 369--386
  (2019)

\bibitem{DBLP:conf/bmvc/SuM17}
Su, J., Maji, S.: Adapting models to signal degradation using distillation. In:
  British Machine Vision Conference (2017)

\bibitem{DBLP:conf/cvpr/SzegedyVISW16}
Szegedy, C., Vanhoucke, V., Ioffe, S., Shlens, J., Wojna, Z.: Rethinking the
  inception architecture for computer vision. In: {IEEE/CVF} Conference on
  Computer Vision and Pattern Recognition. pp. 2818--2826 (2016)

\bibitem{tian2019contrastive}
Tian, Y., Krishnan, D., Isola, P.: Contrastive representation distillation. In:
  International Conference on Learning Representations (2020)

\bibitem{urban2016deep}
Urban, G., Geras, K.J., Kahou, S.E., Aslan, {\"{O}}., Wang, S., Mohamed, A.,
  Philipose, M., Richardson, M., Caruana, R.: Do deep convolutional nets really
  need to be deep and convolutional? In: International Conference on Learning
  Representations (2017)

\bibitem{DBLP:journals/pami/WangY22}
Wang, L., Yoon, K.: Knowledge distillation and student-teacher learning for
  visual intelligence: {A} review and new outlooks. {IEEE} Trans. Pattern Anal.
  Mach. Intell.  \textbf{44}(6),  3048--3068 (2022)

\bibitem{xie2020self}
Xie, Q., Luong, M.T., Hovy, E., Le, Q.V.: Self-training with noisy student
  improves imagenet classification. In: {IEEE/CVF} Conference on Computer
  Vision and Pattern Recognition. pp. 10687--10698 (2020)

\bibitem{DBLP:conf/cvpr/YuanTLWF20}
Yuan, L., Tay, F.E.H., Li, G., Wang, T., Feng, J.: Revisiting knowledge
  distillation via label smoothing regularization. In: {IEEE/CVF} Conference on
  Computer Vision and Pattern Recognition. pp. 3902--3910 (2020)

\bibitem{zhang2018deep}
Zhang, Y., Xiang, T., Hospedales, T.M., Lu, H.: Deep mutual learning. In:
  {IEEE/CVF} Conference on Computer Vision and Pattern Recognition. pp.
  4320--4328 (2018)

\end{thebibliography}
\end{document}